\documentclass{ceurart}

\usepackage{amsmath}
\usepackage{amssymb}
\usepackage{booktabs}
\usepackage{tabularx}
\usepackage{longtable}
\usepackage{array}
\usepackage{hyperref}

\begin{document}

\copyrightyear{2026}
\copyrightclause{Copyright for this paper by its authors.
  Use permitted under Creative Commons License Attribution 4.0
  International (CC BY 4.0).}

\conference{LLM4XAI 2026: Generative Models for Explainable AI Narratives,
  at CIKM 2026, November 8, 2026, Rome, Italy}

\title{Reasoning Externalization for Faithful Large Language Model Narratives of Stock Return Predictions}

\author[1]{Sujung Kim}[
email=sujung2026@hanyang.ac.kr,
]
\fnmark[1]

\author[1]{Seung Hwan Cho}[
email=shcho95@hanyang.ac.kr,
]
\fnmark[1]

\author[1,2]{Sangjin Park}[
orcid=0000-0002-6055-0376,
email=psj3493@hanyang.ac.kr,
]
\cormark[1]

\author[1,2]{Young-Min Kim}[
email=yngmnkim@hanyang.ac.kr,
]

\address[1]{Department of Industrial Data Engineering, Hanyang University, Republic of Korea}
\address[2]{School of Interdisciplinary Industrial Studies, Hanyang University, Republic of Korea}

\fntext[1]{These authors contributed equally to this work.}
\cortext[1]{Corresponding author.}

\begin{abstract}
In finance, interpreting machine learning predictions is essential, yet the numerical outputs of explainable AI can be difficult for non-experts to understand. 
While large language models (LLMs) can translate these outputs into natural language, they may produce errors when inferring numerical changes and feature relations. 
We propose an LLM narrative framework for cross-sectional stock return prediction that combines temporal Shapley additive explanations (SHAP) evidence with historical regime analogs. 
Temporal evidence tracks changes in the normalized global SHAP importance of an XGBoost model over six months. 
Historical analogs are past periods with similar changes in SHAP importance, their model performance and subsequent market returns are provided as comparative context. 
Using this framework, we conduct a controlled study of progressive reasoning externalization, sequentially providing raw SHAP sequences, deterministic temporal descriptors, and feature relations.  
Each generated claim is verified against provenance-linked evidence.  
Across Qwen3, externalizing numerical and relational reasoning improved evidence faithfulness as well as temporal and relational accuracy.  
Evidence faithfulness increased from 0.696 to 0.996 for Qwen3-32B-Instruct. 
While historical analogs did not improve structured automatic faithfulness, they received higher human-rated usefulness scores. 
These results suggest that externalizing verifiable reasoning enhances narrative faithfulness and that historical context adds interpretive value. 
The code is available at \url{https://github.com/sugenre/reasoning-externalization-xai}.
\end{abstract}


\maketitle

\section{Introduction}
Machine learning based predictive models are increasingly being adopted in the financial domain; however, their inherently black box nature makes it difficult to understand the rationale underlying their predictions. 
Model interpretability is considered essential in finance for both investment decision making and regulatory compliance \cite{Baviskar2025}, motivating the adoption of explainable AI (XAI) methods. 
Among the most widely used XAI methods, Shapley additive explanations (SHAP) quantitatively measure the contribution of individual features to model predictions \cite{Lundberg:2017aa}. 
However, SHAP outputs derived from complex financial data are inherently numerical and can be difficult for non expert users to understand \cite{Zeng:2024aa}.
Translating such quantitative explanations into insights that can support decision making therefore remains a challenging task \cite{Zytek:2024aa}.

To address this limitation, recent studies have used large language models (LLMs) to translate SHAP outputs into natural language explanations, improving understandability \cite{Martens:2025aa}. 
However, return-generating processes in financial markets vary over time due to business cycles, policy changes, crises, and other structural shifts \cite{Jenett:2026aa}. 
Therefore, a SHAP explanation based on data from a single point in time cannot fully convey how the model interprets current market conditions. 
Historical information on how the model behaved under similar market conditions, and how the market subsequently evolved, can provide valuable context for financial practitioners \cite{Khanna:2025aa}. 
Such temporal and historical context has not yet been systematically incorporated into LLM-based explanations. 
Providing it is itself challenging, as LLMs struggle to reason over time-series data \cite{2025tsver}. 
Narratives generated from this evidence therefore require verification of their faithfulness after generation \cite{Pratama:2026aa}. 
This verification is essential in finance, where model reliability directly affects economic outcomes and regulatory compliance \cite{Baviskar2025}.

In this study, we propose a framework that provides LLMs with temporal XAI evidence to generate financial analysis narratives in a cross-sectional stock return prediction setting. 
An XGBoost model is trained on preprocessed CRSP\footnote{Center for Research in Security Prices (CRSP), accessed via Wharton Research Data Services (WRDS).} data, and its SHAP values are used to construct temporal trajectories of feature attributions. 
Historical periods with similar attribution trajectories are then retrieved and provided as historical context. 
Using this framework, we conduct a controlled study of how progressively externalizing numerical and relational reasoning affects narrative faithfulness. 
The main contributions of this study are as follows.

\begin{itemize}
  \item \textbf{Evidence-grounded narrative framework.} Temporal SHAP evidence is linked to historical regimes with similar changes in SHAP importance, and each generated claim is verified against a provenance-linked evidence registry.
  \item \textbf{Controlled study of reasoning externalization.} Deterministic temporal descriptors and feature relations substantially reduce temporal and relational reasoning errors and improve evidence faithfulness across Qwen3 scales.
  \item \textbf{Role of historical analogs.} Historical analogs do not improve structured automatic faithfulness but increase human-rated usefulness, serving as complementary interpretive context.
\end{itemize}

\section{Related Work}

\subsection{SHAP Based Feature Analysis in Financial Prediction}

SHAP is one of the most widely used XAI methods grounded in game theory \cite{Lundberg:2017aa}. By providing both global attributions for the overall model and local attributions for individual predictions, SHAP has been extensively studied across various domains \cite{Kim:2026aa}, including financial return prediction. Goswami and Uddin analyzed 166 asset-pricing characteristics and found that momentum and trading related features exhibited high contributions, with portfolio analyses further demonstrating their substantial economic significance \cite{Goswami:2026aa}. Wang showed that, in neural network models, momentum- and trading-related features contributed most strongly to the prediction of abnormal stock returns, whereas investor sentiment features exhibited the highest contributions in predicting excess stock returns \cite{Wang:2024aa}. However, these studies primarily conducted static SHAP analyses over the full sample period or at specific points in time.

To address the limitations of static analysis, Lundberg et al. proposed a method that uses SHAP to quantify each feature's contribution to predictive loss and tracks these contributions over time to identify the causes of model performance degradation \cite{Lundberg:2020aa}. Rather than directly comparing shifts in input distributions, Mougan et al. introduced the concept of Explanation Shift, which detects changes in how a model utilizes features by comparing historical and current SHAP based distributions \cite{Mougan:2023aa}. Similarly, Jenett et al. applied XGBoost and SHAP to the prediction of Real Estate Investment Trust returns and volatility, showing that the importance of key variables changes across distinct market regimes, including the global financial crisis, the low interest rate period, and the COVID-19 pandemic. They further analyzed nonlinear relationships between explanatory variables and returns using Accumulated Local Effects \cite{Jenett:2026aa}. These studies examined dynamic changes in SHAP importance primarily from the perspective of monitoring and detecting model changes, but did not exploit temporal patterns in these changes to identify historically similar market regimes or support downstream analysis.

\subsection{LLM-Based Explanation and Context-Augmented Financial Analysis}

Although SHAP based outputs provide detailed model explanations, they can be difficult for non expert users to understand and interpret. 
To address this issue, Zeng and Zhu proposed a pipeline that organizes SHAP outputs into a structured format and generates natural language explanations through prompt engineering, thereby improving the clarity and usability of model explanations \cite{Zeng:2024aa}. 
Zytek et al. introduced Explingo, which consists of a Narrator that translates ML explanations into natural language and a Grader that evaluates the generated explanations, with the aim of producing high quality narratives \cite{Zytek:2024aa}. 
Martens et al. proposed XAIstories, which transforms SHAP and counterfactual explanations into LLM generated narratives, and showed that these narratives helped users summarize and understand AI decisions more accurately than raw SHAP outputs \cite{Martens:2025aa}. 
Geng et al. found that providing SHAP based feature rankings to an LLM resulted in better performance than allowing the LLM to infer feature importance independently, suggesting that LLMs should be used as controlled narrative interfaces \cite{Geng:2026aa}. 
Beyond SHAP, Wang translated explanations of a temporal graph convolutional network for stock trend prediction into natural language financial reports using LLMs, and evaluated the reports with a factual sensitivity protocol \cite{Wang:2025aa}. 
Nevertheless, concerns remain regarding the faithfulness of generated narratives. 
Lukassen et al. pointed out that prior studies have predominantly evaluated the textual quality of generated narratives while providing limited validation of their practical usefulness for decision making \cite{Lukassen:2026aa}. 
Pratama and Tseng further identified fidelity failures in LLM generated credit risk reports, including reversals of SHAP value signs, omission of dominant features, and inclusion of features that were not provided in the underlying evidence \cite{Pratama:2026aa}.

A growing body of research has also investigated the use of contextual information in financial analysis and prediction. Teixeira et al. proposed Labeled Guide Prompting, which combines structured outputs from Bayesian Networks with LLMs to automatically generate credit risk reports \cite{Teixeira:2023aa}. Kim et al. provided GPT-4 with financial statements to predict the direction of future earnings changes and demonstrated that LLMs can generate useful narrative insights regarding future earnings \cite{Kim:2024aa}. Fatouros et al. introduced MarketSenseAI 2.0, which integrates news, financial statements, and macroeconomic data within a multiagent LLM architecture to support stock analysis and investment decision making \cite{Fatouros:2025aa}. These studies primarily used textual or numerical information observed at a given point in time as contextual input to LLMs. Other studies have explored the use of historically similar cases as contextual information. Khanna et al. proposed a framework that combines macroeconomic indicators with text embeddings to retrieve similar historical periods and provides the retrieved cases as context for LLM based prediction \cite{Khanna:2025aa}. In their framework, similarity is measured based on macroeconomic variables, while temporal patterns in feature attributions derived from an ML model are not used.

\section{Methodology}

\begin{figure}[h!]
  \centering
  \includegraphics[width=\linewidth]{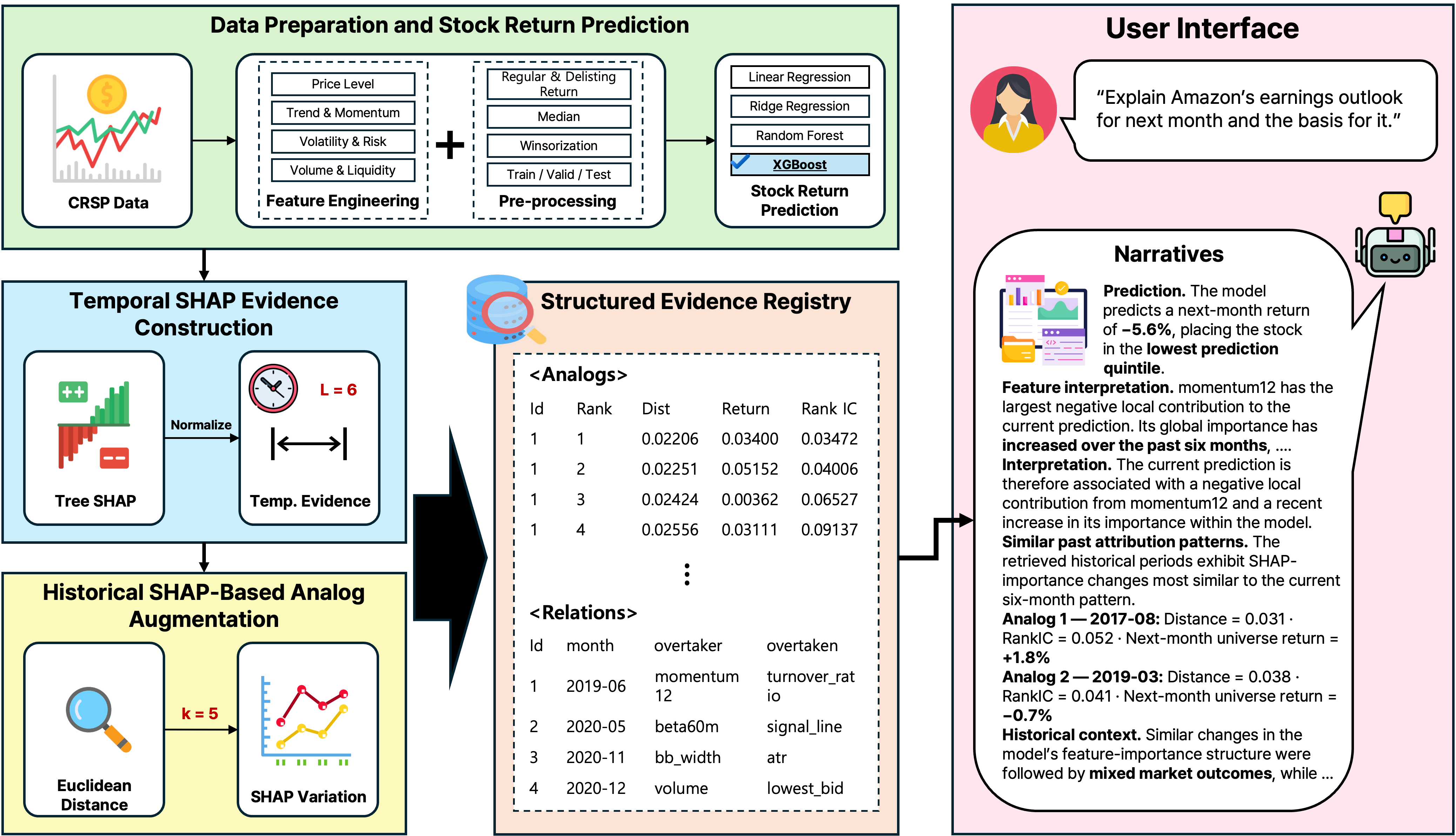}
  \caption{Overview of the proposed framework for generating evidence-grounded LLM narratives of stock return predictions.}
  \label{fig:framework}
\end{figure}

\subsection{Data Preparation and Stock Return Prediction}

Using monthly CRSP data obtained through WRDS, we constructed 30 stock level characteristics covering price and trend, momentum, volatility and risk, and trading volume and liquidity. Each characteristic was defined as either a month end value or a monthly aggregate according to its economic interpretation and the measurement convention of the underlying raw data. For delisting observations, both regular returns and delisting returns were incorporated. Missing characteristic values were imputed using the monthly cross sectional median, after which the remaining missing values were set to zero following \cite{Gu2020}. To mitigate the influence of extreme observations while maintaining consistent preprocessing across time, each characteristic was winsorized using the 1st and 99th percentile thresholds estimated from the training sample, and the same thresholds were subsequently applied to the validation and test periods \cite{Gray2024}. Preserving the temporal ordering of the data, the sample was divided into a training period from January 1995 to December 2009, a validation period from January 2010 to December 2011, and a test period from January 2012 to November 2023.

Using the resulting monthly stock characteristics, we formulated a cross sectional stock return prediction task in which the feature vector $\mathbf{x}_{i,t}$ of stock $i$ observed in month $t$ is used to predict its realized return $r_{i,t+1}$ in the following month. Monthly prediction is a widely adopted setting in cross sectional asset pricing research based on firm characteristics \cite{Gu2020} and is also well suited to aligning characteristics with heterogeneous update frequencies at a common time point. Because the primary focus of this study is not short term price fluctuations themselves but rather the model's feature dependence structure and its temporal variation, both the return prediction task and the subsequent SHAP based temporal analysis were conducted at a monthly frequency.

\subsection{Temporal XAI Evidence Construction}

Tree based models can flexibly capture nonlinear relationships and interactions among features in cross sectional return prediction \cite{Gu2020}, while TreeSHAP enables efficient computation of feature attributions for individual predictions \cite{Lundberg:2020aa}. Let the SHAP value for stock $i$, feature $j$, and month $t$ be denoted by $\phi_{i,j,t}$. The monthly global attribution was computed by first taking the cross sectional mean of the absolute SHAP values for each feature and then normalizing it by the sum of these mean absolute attributions across all features to obtain $P_{j,t}$. The resulting $P_{j,t}$ represents the relative importance of feature $j$ in the model's global attribution for month $t$. Let $P_t$ denote the 30 dimensional global attribution vector for month $t$; the temporal evidence over the most recent six months was then constructed as $S_t$.

\begin{align}
A_{j,t}
&= \frac{1}{N_t}\sum_i \left|\phi_{i,j,t}\right|,
\label{eq:mean-absolute-shap}\\
P_{j,t}
&= \frac{A_{j,t}}{\sum_k A_{k,t}},
\label{eq:normalized-global-importance}\\
S_t
&= \left[P_{t-5},P_{t-4},\ldots,P_t\right] \in \mathbb{R}^{6\times 30}.
\label{eq:temporal-evidence}
\end{align}

\subsection{Historical SHAP Based Analog Augmentation}

To augment temporal XAI evidence with historical comparative context, we retrieved past market regimes exhibiting similar patterns of change in feature importance. While raw market variables such as prices and returns can characterize similarity in market states, changes in global SHAP importance reflect how the relative contribution structure of features evolves over time under a fixed predictive model. We therefore retrieved historical regimes with similar SHAP importance dynamics and provided the model's predictive performance and subsequent universe returns in those periods as comparative context.

Using $P_t$, we defined the monthly change in attribution as in Equation~\eqref{eq:monthly-shap-change} and used it as the retrieval representation. For a temporal window of length $L$, the query trajectory at month $t$ was constructed as in Equation~\eqref{eq:analog-query-trajectory}. Because each $\Delta P_t$ is a 30 dimensional vector, $\Delta P_t \in \mathbb{R}^{30}$, the distance between query trajectory $Q_t^{(L)}$ and historical candidate trajectory $Q_s^{(L)}$ was computed using Euclidean distance.

\begin{align}
\Delta P_t
&= P_t-P_{t-1},
\label{eq:monthly-shap-change}\\
Q_t^{(L)}
&= \left[\Delta P_{t-L+1},\ldots,\Delta P_t\right],
\label{eq:analog-query-trajectory}\\
d(t,s)
&= \left\|\operatorname{vec}\left(Q_t^{(L)}\right)
-\operatorname{vec}\left(Q_s^{(L)}\right)\right\|_2.
\label{eq:analog-distance}
\end{align}

To ensure that the query and candidate trajectories did not overlap and that the outcome following each candidate was observable before the query period, historical candidates were restricted to $s \leq t-(L+1)$. The top $K$ candidates with the smallest distances were selected as historical analogs. The sensitivity to $L$ and $K$, as well as the final retrieval configuration, was determined empirically.

Each historical analog was associated with its trajectory distance, monthly RankIC at the corresponding period, and the equal weighted universe return in the subsequent month. These outcomes were not treated as direct forecasts or causal evidence. Rather, they served as comparative historical context indicating the model's predictive performance and subsequent universe returns during past periods exhibiting similar changes in feature attribution.

\subsection{Structured Evidence Registry}

The generated prediction, feature attribution, temporal descriptors, feature relations, and historical analog evidence were stored in a unified structured evidence registry. Each evidence unit was assigned a unique provenance identifier specifying the case, evidence type, and associated feature or relation. For example, \texttt{case\_id\#TMP:momentum12} denotes the temporal numerical evidence for \texttt{momentum12} in a given case.

All experimental conditions shared the same underlying registry, while the visibility of each evidence unit was predetermined for each condition. Thus, differences across conditions were created by varying only the scope of information exposed to the LLM, rather than by recomputing the underlying evidence. The prompt presented each evidence item together with its corresponding identifier, which was subsequently used to verify generated claims against the source evidence.

\subsection{Reasoning Externalization and Narrative Generation}

To control the reasoning burden placed on the LLM during temporal XAI narrative generation, numerical and relational reasoning were progressively externalized. 
Motivated by findings that delegating computation from the LLM to an external interpreter improves reasoning accuracy~\cite{pal2023}, these computations were performed deterministically before generation and provided as evidence rather than derived by the LLM. 
The experimental conditions were designed according to the dependency structure of the required reasoning steps, while the system prompt and output schema were kept unchanged across conditions.

\begin{table}[h!]
  \centering
  \caption{Experimental Conditions and Evidence Provided}
  \label{tab:experimental-conditions}
  \begin{tabularx}{\linewidth}{@{}lX@{}}
    \toprule
    \textbf{Condition} & \textbf{Evidence provided} \\
    \midrule
    C0-Evidence Control & Prediction, local SHAP, and current global attribution \\
    C1-Raw & C0 + raw 6-month global attribution sequence \\
    C2-Numerical & C1 + $\Delta6P$, slope, and trend direction \\
    C3-Relational & C2 + deterministic OVERTAKES relations \\
    C3-Analog & C3 + historical analog evidence \\
    \bottomrule
  \end{tabularx}
\end{table}

C0 evaluates whether the model suppresses unsupported temporal inference when no temporal evidence is available. 
Under C1, the LLM must infer temporal trends and feature relations directly from the raw temporal sequence. 
C2 externalizes deterministic numerical computation by explicitly providing temporal statistics. 
The six-month attribution change was defined as 
$\Delta_6P_{j,t}=P_{j,t}-P_{j,t-5}$. 
The temporal slope was estimated by OLS over the six monthly attribution values 
$P_{j,t-5:t}$, using month indices $0,\ldots,5$. 
Trend direction was labeled as increasing, decreasing, or flat according to the sign of the six month OLS slope. 
C3 further provides OVERTAKES relations, thereby removing the burden of relational inference from the LLM. 
Feature $a$ was defined to overtake feature $b$ when 
$P_{a,t-5}<P_{b,t-5}$ but $P_{a,t}>P_{b,t}$. 
C3-Analog is a separate augmentation condition that combines C3 with historical context. 
Each output consisted of natural language text together with structured claims. 
Every factual claim was required to reference evidence IDs available under the corresponding condition, while the model was allowed to abstain when no supporting evidence was available.

The primary model was Qwen3-32B-Instruct \cite{Yang2025Qwen3}, the largest dense model that could be executed under a consistent inference protocol in our experimental environment. Qwen3-14B and Qwen3-8B were additionally included to assess whether the observed effects remained consistent across model scales within the same model family, while their corresponding Base checkpoints were used to examine sensitivity to post training. Ministral-8B-Instruct-2410 \cite{MistralAI2024Ministral8B} and Llama-3.1-8B-Instruct \cite{Grattafiori2024Llama3} were further included to evaluate whether the effects of reasoning externalization remained consistent across different model families.

\section{Experiments and Results}

\subsection{Evaluation Protocol}

\subsubsection{Prediction Performance Evaluation}

Return prediction performance was evaluated using RankIC, RankICIR, RMSE, and MAE. For each month $t$, RankIC was computed as the cross sectional Spearman correlation between the predicted next-month return $\widehat{r}_{i,t+1}$ and the realized next month return $r_{i,t+1}$ across stocks. We report the average monthly RankIC over the full test period. RankICIR was calculated as the mean monthly RankIC divided by its standard deviation, measuring the stability of cross-sectional ranking performance over time.

\begin{align}
\operatorname{RankIC}_t
&= \operatorname{Spearman}_i\!\left(\widehat{r}_{i,t+1}, r_{i,t+1}\right),
\label{eq:monthly-rankic}\\
\operatorname{RankIC}
&= \frac{1}{T}\sum_{t=1}^{T}\operatorname{RankIC}_t,
\label{eq:mean-rankic}\\
\operatorname{RankICIR}
&= \frac{\operatorname{mean}(\operatorname{RankIC}_t)}
{\operatorname{std}(\operatorname{RankIC}_t)}.
\label{eq:rankicir}
\end{align}

\subsubsection{Deterministic Automatic Evaluation}

The generated outputs were evaluated by directly matching their structured semantic fields against the evidence registry. Our primary metric, Structured Evidence Faithfulness (EF), was defined as the proportion of factual claims that referenced evidence permitted under the corresponding experimental condition and matched the ground truth subject--predicate--value tuple.

Because the evaluation operates on structured fields generated alongside the natural language narrative, EF measures the consistency between structured factual claims and their supporting evidence rather than the overall semantic truthfulness of unrestricted free form text. Output format reliability was additionally evaluated using the parse rate. For C0, in which no temporal evidence was provided, we further evaluated whether the model appropriately refrained from unsupported inference using Evidence Control Abstention Accuracy.

\subsubsection{LLM-as-a-Judge}

Because the deterministic evaluation verifies only the correspondence between structured claim fields and the evidence registry, without directly assessing the narrative text itself, we conducted a complementary LLM-as-a-Judge evaluation on the C3-Relational and C3-Analog conditions. Using month stratified sampling, we selected a common set of 400 \texttt{case\_id}s and paired the narratives from the two conditions at the case level. Claude Opus 5, which belongs to a different model family from the generation models (Qwen, Llama, and Mistral), was used as the judge. Model and condition identifiers were omitted from the evaluation prompts. Each item was evaluated once ($k=1$) through the Message Batches API.

The evaluation comprised three criteria, with different levels of information exposure for each criterion. First, the judge was shown only the narrative text, without any supporting evidence, and asked to determine whether the normalized global SHAP importance ($P$) of the focal feature had increased, decreased, or remained unchanged over the preceding six months, or whether the direction could not be determined from the text alone. Textual Temporal Transfer Accuracy was computed as the exact match
accuracy between the resulting judgment and \texttt{trend\_direction} in the evidence registry.

For the second and third criteria, the judge was provided with both the narrative and the evidence block corresponding to the relevant condition. Evidence Scope Adherence was defined as the extent to which the narrative avoided claims beyond the scope of the provided evidence, whereas Narrative Synthesis was defined as the extent to which multiple evidence items were coherently integrated into an interpretation while remaining within the supported evidence scope. Both criteria were evaluated using pre defined anchored 1--5 rating rubrics.

\subsubsection{Human Evaluation}

To assess the practical utility of historical analog context, we conducted a human evaluation comparing narratives generated by Qwen3-32B-Instruct under the C3-Relational and C3-Analog conditions. Ten evaluators each assessed two cases. For each case, the evaluator reviewed one C3-Relational narrative and one C3-Analog narrative, resulting in four narrative evaluations per evaluator. Evaluators were non expert users with an interest in stock investment and basic familiarity with artificial intelligence concepts. Condition labels were hidden, and both the case order and the order of the two narratives within each pair were randomized. Evaluators rated each narrative using five criteria: Understandability, Decision Usefulness, Contextual Usefulness, Reliance Calibration, and Adoption Intention, on five point Likert scales. This evaluation was designed to examine whether historical analogs provide additional value for users’ interpretation and use of the narratives, separately from automatic faithfulness evaluation.

\subsection{Experimental Setup and Prediction Model Selection}

All experiments were conducted on Ubuntu 24.04 using an NVIDIA RTX PRO 6000 GPU with 96 GB of memory. XGBoost 3.2.0 and SHAP 0.51.0 were used for prediction and XAI analysis, respectively, while LLM inference was performed using vLLM 0.11.2. The same decoding configuration was applied across all LLM conditions.

\begin{table}[h!]
  \centering
  \small
  \caption{Out-of-Sample Prediction Performance}
  \label{tab:prediction-performance}
  \begin{tabular}{@{}lrrrr@{}}
    \toprule
    \textbf{Model} 
    & \textbf{RankIC $\uparrow$} 
    & \textbf{RankICIR $\uparrow$} 
    & \textbf{RMSE $\downarrow$} 
    & \textbf{MAE $\downarrow$} \\
    \midrule
    
    Linear Regression 
    & 0.0170 
    & 0.2380 
    & \underline{0.1631} 
    & \underline{0.0845} \\
    
    Ridge Regression 
    & 0.0209 
    & 0.2810 
    & \underline{0.1631} 
    & 0.0848 \\
    
    Random Forest 
    & \underline{0.0357} 
    & \underline{0.4900} 
    & \textbf{0.1627} 
    & \textbf{0.0841} \\
    
    XGBoost 
    & \textbf{0.0361} 
    & \textbf{0.5380} 
    & 0.1660 
    & 0.0859 \\
    
    \bottomrule
  \end{tabular}
\end{table}

Because the primary objective was not to compete for state-of-the-art predictive performance, but rather to evaluate the faithfulness of LLM-generated narratives grounded in XAI evidence derived from a fixed prediction model, we compared two linear baselines, Linear Regression and Ridge Regression, with two tree-based models, Random Forest and XGBoost, which can capture nonlinear relationships and feature interactions while supporting TreeSHAP-based attribution.

The tree based models achieved higher RankIC values than the linear models, suggesting that capturing nonlinear effects and feature interactions may be important for modeling the relationship between financial characteristics and next month returns in this prediction setting. Compared with Random Forest, XGBoost achieved 1.4\% higher RankIC and 9.8\% higher RankICIR. However, Random Forest achieved lower RMSE and MAE, indicating that XGBoost did not dominate across all predictive metrics. XGBoost was selected using validation period performance and then fixed for all test period XAI analyses. Test results were used only for final out of sample evaluation. TreeSHAP was applied to interpret feature contributions under this predictive model.

\subsection{Historical Analog Characterization}

Before incorporating historical analogs as contextual evidence, we assessed
whether regimes retrieved based on changes in SHAP importance provided useful
comparative context for subsequent universe returns. Sensitivity analyses were
conducted across temporal window lengths $L \in \{3,6,9\}$ and numbers of
retrieved analogs $K \in \{1,3,5\}$.

For comparison, we constructed a market state based retrieval method that
retained the same retrieval conditions but replaced the representation with
changes in the cross sectional market state derived from the raw variables.
We additionally evaluated a recent return baseline and a random baseline that
selected historical regimes at random. For each method, we compared the
aggregated subsequent universe returns of the retrieved historical regimes
with the realized next month universe return using mean absolute error (MAE).
Table~\ref{tab:analog-sensitivity} reports the results across the evaluated
retrieval configurations.

\begin{table}[htbp]
\centering
\small
\caption{Mean Absolute Error of Subsequent Universe Returns across Retrieval Settings}
\label{tab:analog-sensitivity}
\setlength{\tabcolsep}{3pt}

\begin{tabular}{@{}lccccccccc@{}}
\toprule
\textbf{Method}
& \multicolumn{3}{c}{\textbf{$K=1$}}
& \multicolumn{3}{c}{\textbf{$K=3$}}
& \multicolumn{3}{c}{\textbf{$K=5$}} \\
\cmidrule(lr){2-4}
\cmidrule(lr){5-7}
\cmidrule(lr){8-10}

& $L=3$ & $L=6$ & $L=9$
& $L=3$ & $L=6$ & $L=9$
& $L=3$ & $L=6$ & $L=9$ \\
\midrule

SHAP-Trajectory
& 0.05146
& 0.05135
& \underline{0.04845}
& \textbf{0.04180}
& 0.04260
& 0.04091
& \textbf{0.04012}
& \textbf{0.03996}
& 0.04121 \\

Market-State
& \underline{0.04752}
& \underline{0.04822}
& 0.04899
& \underline{0.04225}
& \textbf{0.04076}
& \underline{0.04057}
& 0.04098
& \underline{0.03997}
& 0.04147 \\

Recent-Return
& \textbf{0.04544}
& \textbf{0.04128}
& \textbf{0.03972}
& 0.04544
& \underline{0.04128}
& \textbf{0.03972}
& 0.04544
& 0.04128
& \textbf{0.04042} \\

Random
& 0.04902
& 0.04858
& 0.04874
& 0.04236
& 0.04205
& 0.04186
& \underline{0.04047}
& 0.04040
& \underline{0.04114} \\

\bottomrule
\end{tabular}

\end{table}

For SHAP-Trajectory retrieval, MAE generally decreased as the number of retrieved analogs increased. This pattern is consistent with the possibility that aggregating multiple historical analogs reduces sensitivity to any single retrieved period. Among the evaluated SHAP-Trajectory configurations, $K=5$ and $L=6$ minimized validation MAE, yielding a value of 0.03996. Accordingly, we selected this configuration for SHAP-Trajectory retrieval before evaluating C3-Analog on the test period.

However, SHAP-Trajectory retrieval was not consistently superior to the
alternative methods. The recent return baseline achieved lower MAE than
SHAP-Trajectory in six of the nine evaluated configurations. Under the selected $K=5$ and $L=6$ configuration, SHAP-Trajectory achieved an MAE of 0.03996,
which was nearly identical to the 0.03997 obtained by market-state retrieval.
We therefore do not interpret these results as evidence of consistent
superiority of SHAP-Trajectory retrieval over the alternative methods.
Instead, the retrieved analogs are used as historical comparative context,
providing information on model performance and subsequent equal-weighted
universe returns during past periods characterized by similar changes in model
feature contributions.

\subsection{Reasoning Externalization and Narrative Faithfulness}

Table~\ref{tab:cross-model-consistency} reports cross-model performance under C1--C3 in terms of evidence faithfulness, temporal accuracy, and relational accuracy, together with output parse rates.

\suppressfloats[t]

\begin{table}[htbp]
\centering
\small
\caption{Cross-Model Consistency of Reasoning Externalization}
\label{tab:cross-model-consistency}
\resizebox{\linewidth}{!}{%
\begin{tabular}{@{}lcccccccccc@{}}
\toprule
\textbf{Model}
& \textbf{Parse}
& \textbf{EF C1}
& \textbf{EF C2}
& \textbf{EF C3}
& \textbf{Temp. C1}
& \textbf{Temp. C2}
& \textbf{Temp. C3}
& \textbf{Rel. C1}
& \textbf{Rel. C2}
& \textbf{Rel. C3} \\
\midrule

Qwen3-8B
& 0.9363
& 0.7219
& 0.7662
& 0.9195
& \underline{0.6318}
& 0.8784
& 0.8458
& 0.2155
& 0.3585
& 0.9295 \\

Qwen3-8B-Instruct
& \textbf{1.0000}
& \underline{0.8259}
& \textbf{0.9704}
& 0.9686
& 0.3604
& 0.6125
& 0.5750
& \textbf{0.8625}
& \textbf{0.8875}
& \textbf{1.0000} \\

Qwen3-14B
& 0.9779
& 0.7570
& 0.8061
& \underline{0.9984}
& \textbf{0.6979}
& \textbf{0.9916}
& \textbf{0.9916}
& 0.3375
& 0.2059
& \underline{0.9979} \\

Qwen3-14B-Instruct
& \textbf{1.0000}
& 0.7194
& 0.8844
& \textbf{0.9995}
& 0.4854
& 0.9479
& \underline{0.9521}
& 0.4271
& 0.4979
& \textbf{1.0000} \\

Qwen3-32B-Instruct
& \underline{0.9988}
& 0.6960
& \underline{0.9032}
& 0.9964
& 0.3396
& \underline{0.9896}
& 0.9313
& 0.5250
& 0.5729
& \textbf{1.0000} \\

Ministral-8B-Instruct
& 0.5433
& \textbf{0.8733}
& 0.8536
& 0.9204
& 0.5506
& 0.6273
& 0.5246
& \underline{0.7341}
& \underline{0.8636}
& 0.9836 \\

Llama-3.1-8B-Instruct
& 0.4525
& 0.6555
& 0.5762
& 0.5437
& 0.1905
& 0.4020
& 0.4098
& 0.5450
& 0.4975
& 0.7639 \\

\bottomrule
\end{tabular}%
}
\end{table}

Across all Qwen3 Instruct models, numerical externalization from C1 to C2 increased Structured Temporal Accuracy, with improvements of 0.6500, 0.4625, and 0.2521 for the 32B, 14B, and 8B models, respectively. When relational evidence was provided in C3, Relational Accuracy increased over C2 by 0.4271, 0.5021, and 0.1125, respectively, showing the same directional effect regardless of model scale. Ministral and Llama exhibited relatively low parse rates of 0.5433 and 0.4525, respectively, resulting in different numbers of valid samples across conditions; they were therefore excluded from direct quantitative comparisons. The Qwen3 base models also showed increases in Structured Temporal Accuracy from C1 to C2 of 0.2466 for 8B and 0.2937 for 14B, indicating that the effect of numerical externalization was not limited to instruction-tuned models. Meanwhile, Qwen3 Base models generally achieved higher temporal reasoning accuracy than their Instruct counterparts of the same size, but lower relational reasoning accuracy, suggesting that the effect of instruction tuning may differ across reasoning types.

\begin{table}[htbp]
\centering
\small
\caption{Structured Claim Performance by Reasoning Condition (Qwen3-32B-Instruct)}
\label{tab:qwen32-structured-claim}
\begin{tabular}{@{}lcccc@{}}
\toprule
\textbf{Condition}
& \textbf{EF $\uparrow$}
& \textbf{Temporal Acc. $\uparrow$}
& \textbf{Relational Acc. $\uparrow$}
& \textbf{C0 Abstention} \\
\midrule
C0-Evidence Control
& 0.9104 & -- & -- & 0.8386 \\
C1-Raw
& 0.6960 & 0.3396 & 0.5250 & -- \\
C2-Numerical
& 0.9032 & \textbf{0.9896} & \underline{0.5729} & -- \\
C3-Relational
& \textbf{0.9964} & \underline{0.9313} & \textbf{1.0000} & -- \\
C3-Analog
& \underline{0.9891} & 0.9000 & \textbf{1.0000} & -- \\
\bottomrule
\end{tabular}
\end{table}

For Qwen3-32B-Instruct, externalizing numerical information from C1 to C2 increased EF from 0.6960 to 0.9032, corresponding to an improvement of 0.2072, while Structured Temporal Accuracy increased from 0.3396 to 0.9896, an improvement of 0.6500. This substantial increase indicates that explicitly providing deterministic temporal descriptors markedly improved the model's ability to represent temporal changes relative to requiring such information to be inferred from raw attribution sequences. When relational evidence was additionally structured and provided in C3, Relational Accuracy increased from 0.5729 under C2 to 1.0000, corresponding to an improvement of 0.4271. EF also increased from 0.9032 to 0.9964, while Structured Temporal Accuracy remained high at 0.9313. Under the C0-Evidence Control condition, abstention accuracy was 0.8386, indicating that unsupported inference was frequently, though not perfectly, suppressed when temporal evidence was unavailable. C3-Analog maintained perfect Relational Accuracy at 1.0000, while EF and Structured Temporal Accuracy were slightly lower than those under C3-Relational, decreasing from 0.9964 to 0.9891 and from 0.9313 to 0.9000, respectively. These results suggest that historical analog augmentation did not provide an additional improvement in structured automatic faithfulness, motivating a separate evaluation of its narrative level and human level utility.

\subsection{LLM-as-a-Judge Evaluation}

To assess narrative-level faithfulness, we compare C3-Relational
and C3-Analog in terms of temporal information transfer,
evidence-scope adherence, and narrative synthesis.
Table~\ref{tab:llm-judge-evaluation} summarizes the results.

\begin{table}[htbp]
\centering
\small
\caption{LLM-as-a-Judge Evaluation of Narrative Level Faithfulness 
(Mean $\pm$ SD)}
\label{tab:llm-judge-evaluation}
\begin{tabular}{@{}lcc@{}}
\toprule
\textbf{Metric}
& \textbf{C3-Relational}
& \textbf{C3-Analog} \\
\midrule
Textual Temporal Transfer Accuracy
& \textbf{0.9975}
& \textbf{0.9975} \\

Deterministic Temporal Accuracy
& \textbf{0.8625}
& \underline{0.7425} \\

Evidence Scope Adherence
& $\mathbf{4.9300} \pm 0.2555$
& $\underline{4.8900} \pm 0.3212$ \\

Narrative Synthesis
& $\mathbf{2.0275} \pm 0.1637$
& $\underline{1.9975} \pm 0.0867$ \\
\bottomrule
\end{tabular}

\end{table}

Textual Temporal Transfer Accuracy was 0.9975 (399/400) for both conditions, indicating that the addition of historical analogs did not materially impair the communication of temporal change information for the focal feature. In contrast, Deterministic Temporal Accuracy was lower, at 0.8625 for C3-Relational and 0.7425 for C3-Analog. Further analysis showed that all 158 cases classified as failures by the deterministic metric lacked a \texttt{temporal\_trend} claim but contained a \texttt{temporal\_change} claim. This discrepancy therefore reflects a difference in claim-type-specific aggregation rather than an actual loss of temporal information in the narrative.

Evidence Scope Adherence remained high in both conditions, at approximately 4.9 out of 5, suggesting that the inclusion of historical analogs did not increase the generation of claims beyond the scope of the provided evidence. By contrast, Narrative Synthesis scores were low, at approximately 2 out of 5. This suggests that evidence constrained generation is effective in maintaining faithfulness to the provided evidence, but may limit the flexibility with which multiple evidence items are integrated into a narrative.

\subsection{Human Evaluation of Historical Analog Context}

To assess whether historical analogs improve the perceived usefulness
of generated narratives, we compare C3-Relational and C3-Analog across
five human-evaluation criteria covering understandability, decision
usefulness, contextual usefulness, reliance calibration, and adoption
intention. Table~\ref{tab:human-evaluation} summarizes the mean ratings
and between-condition differences.

\begin{table}[htbp]
\centering
\small
\caption{Human Evaluation of Historical Analog Context (Mean $\pm$ SD)}
\label{tab:human-evaluation}
\begin{tabular}{@{}lccc@{}}
\toprule
\textbf{Item}
& \textbf{C3-Relational}
& \textbf{C3-Analog}
& \textbf{$\Delta$} \\
\midrule
Q1 Understandability
& $\underline{3.2000} \pm 1.2397$
& $\mathbf{3.6000} \pm 1.2732$
& $+0.4000$ \\

Q2 Decision Usefulness
& $\underline{3.3500} \pm 1.2258$
& $\mathbf{3.8000} \pm 1.1517$
& $+0.4500$ \\

Q3 Contextual Usefulness
& $\underline{3.2000} \pm 1.5079$
& $\mathbf{3.7000} \pm 1.2607$
& $+0.5000$ \\

Q4 Reliance Calibration
& $\underline{3.2500} \pm 1.1642$
& $\mathbf{3.6500} \pm 1.1367$
& $+0.4000$ \\

Q5 Adoption Intention
& $\underline{2.8000} \pm 1.3611$
& $\mathbf{3.3500} \pm 1.3870$
& $+0.5500$ \\

\textbf{Overall}
& $\underline{3.1600}$
& $\mathbf{3.6200}$
& $+0.4600$ \\
\bottomrule
\end{tabular}

\end{table}

The human evaluation results showed that C3-Analog achieved higher mean scores than C3-Relational across all evaluation criteria. The overall mean increased from 3.1600 to 3.6200, corresponding to a mean difference of +0.4600. Adoption Intention exhibited the largest difference (+0.5500), followed by Contextual Usefulness (+0.5000). These higher mean ratings were observed even though EF and Structured Temporal Accuracy under C3-Analog were slightly lower than those under C3-Relational in the preceding automatic evaluation. This suggests that historical analog information may provide users with useful comparative context for interpreting and potentially using current predictions, even when it does not further improve structured accuracy. Historical analogs should therefore be interpreted primarily as complementary information that enhances human-level contextual usefulness rather than as a mechanism for improving automatic faithfulness.

\section{Conclusion}

This study examined whether the reliability of financial XAI narratives can deteriorate when LLMs are required to directly derive numerical changes and feature relations from raw XAI evidence, and evaluated an approach that externalizes such reasoning burdens into verifiable evidence. SHAP based temporal evidence was structured into numerical and relational information, while factual claims were linked to Evidence IDs, enabling deterministic evaluation of faithfulness and reasoning accuracy without relying on LLM-as-a-Judge.

The results showed that explicitly providing attribution changes and trend information improved faithfulness and temporal reasoning relative to providing only raw temporal sequences, while externalizing feature relations further improved relational reasoning. These findings suggest that structuring error prone reasoning steps into verifiable evidence can improve the reliability of financial XAI narratives. Future work should examine whether these findings generalize to other predictive models and financial decision making tasks.


\begin{acknowledgments}

This work was supported by the National Research Foundation of
Korea (NRF) grant funded by the Korea government (MSIT)
(No. RS 2025-00554384), and the Technology Development Program
(No. RS 2024-00513926) funded by the Ministry of SMEs and Startups
(MSS, Korea).

\end{acknowledgments}


\bibliography{finanshap}

\appendix

\section{Stock Level Characteristic Definitions}
\label{app:stock-characteristics}

For the price-based characteristics, we define the adjusted price variables as
\[
c_t=
\begin{cases}
1, & \mathrm{CFACPR}_t=0 \text{ or missing},\\
\mathrm{CFACPR}_t, & \text{otherwise}.
\end{cases}
\]

\[
\begin{aligned}
\mathrm{adjp}_t
&= \frac{|\mathrm{PRC}_t|}{c_t},
\qquad
\mathrm{adjhi}_t
= \frac{|\mathrm{ASKHI}_t|}{c_t},
\qquad
\mathrm{adjlo}_t
&= \frac{|\mathrm{BIDLO}_t|}{c_t}.
\end{aligned}
\]

\begingroup

\footnotesize
\setlength{\tabcolsep}{2pt}
\renewcommand{\arraystretch}{1.45}
\setlength{\extrarowheight}{2pt}
\setlength{\LTleft}{0pt}
\setlength{\LTright}{0pt}

\begin{longtable}{
@{}
>{\raggedright\arraybackslash}p{0.24\linewidth}
>{\raggedright\arraybackslash}p{0.44\linewidth}
>{\raggedright\arraybackslash}p{0.26\linewidth}
@{}
}

\caption{Definition and Construction of Stock-Level Characteristics}
\label{tab:stock-characteristics}\\

\toprule
\textbf{Feature}
& \textbf{Definition / Formula}
& \textbf{Measurement \& Source} \\
\midrule
\endfirsthead

\multicolumn{3}{c}{\tablename~\thetable\ (continued)}\\
\toprule
\textbf{Feature}
& \textbf{Definition / Formula}
& \textbf{Measurement \& Source} \\
\midrule
\endhead

\midrule
\multicolumn{3}{r}{Continued on next page}\\
\endfoot

\bottomrule
\endlastfoot


\multicolumn{3}{@{}l}{\textbf{Price Level}}\\
\addlinespace[2pt]

\texttt{lowest\_bid}
& $|\mathrm{BIDLO}_t|$
& Month $t$\par CRSP: BIDLO \\

\texttt{highest\_ask}
& $|\mathrm{ASKHI}_t|$
& Month $t$\par CRSP: ASKHI \\

\texttt{close\_price}
& $|PRC_t|$
& Month $t$\par CRSP: PRC \\

\texttt{ma3}
& $\operatorname{mean}(adjp_{t-2:t})$, minimum 2 observations
& Trailing 3 months\par CRSP: PRC, CFACPR \\

\texttt{ma12}
& $\operatorname{mean}(adjp_{t-11:t})$, minimum 6 observations
& Trailing 12 months\par CRSP: PRC, CFACPR \\

\addlinespace[3pt]


\multicolumn{3}{@{}l}{\textbf{Trend \& Momentum}}\\
\addlinespace[2pt]

\texttt{price\_distance\_ma12}
& $\displaystyle \frac{adjp_t-ma12_t}{ma12_t}$
& Month $t$, relative to 12M MA\par CRSP: PRC, CFACPR \\

\texttt{high\_12m\_ratio}
& $\displaystyle \frac{adjp_t}{\max(adjp_{t-11:t})}$
& Trailing 12 months\par CRSP: PRC, CFACPR \\

\texttt{ma\_cross}
& $+1$ if $ma3_t>ma12_t$; otherwise $-1$
& 3M vs.\ 12M MA\par CRSP: PRC, CFACPR \\

\texttt{momentum12}
& $\displaystyle \frac{adjp_t}{adjp_{t-12}}-1$
& 12-month lag\par CRSP: PRC, CFACPR \\

\texttt{mom1m}
& $RET_{t-1}$
& Previous month\par CRSP: RET \\

\texttt{mom6m}
& $\displaystyle
\exp\!\left(\sum_{k=1}^{6}\log(1+RET_{t-k})\right)-1$
& Previous 6 months\par CRSP: RET \\[5pt]

\texttt{macd}
& $EMA_{12}(adjp)-EMA_{26}(adjp)$
& 12M / 26M EMA\par CRSP: PRC, CFACPR \\

\texttt{signal\_line}
& $EMA_{9}(macd)$
& 9M EMA\par CRSP: PRC, CFACPR \\

\addlinespace[3pt]


\multicolumn{3}{@{}l}{\textbf{Oscillator}}\\
\addlinespace[2pt]

\texttt{stochastic}
& $\displaystyle
100 \times \frac{
  \mathrm{adjp}_t
  - \min_{\tau=t-13,\ldots,t}\mathrm{adjp}_{\tau}
}{
  \max_{\tau=t-13,\ldots,t}\mathrm{adjp}_{\tau}
  - \min_{\tau=t-13,\ldots,t}\mathrm{adjp}_{\tau}
}$
& Trailing 14 months; minimum 6 valid observations
\par CRSP: PRC, CFACPR \\[5pt]

\texttt{rsi}
&
\(
\begin{aligned}
d_t
&= \mathrm{adjp}_t-\mathrm{adjp}_{t-1},\\
g_t
&= \max(d_t,0),\qquad
\ell_t=\max(-d_t,0),\\
G_t
&= \frac{1}{14}\sum_{k=0}^{13}g_{t-k},\\
L_t
&= \frac{1}{14}\sum_{k=0}^{13}\ell_{t-k},\\
RSI_t
&= 100-\frac{100}{1+G_t/L_t}.
\end{aligned}
\)
& Trailing 14 months\par CRSP: PRC, CFACPR \\


\multicolumn{3}{@{}l}{\textbf{Volatility \& Risk}}\\
\addlinespace[2pt]

\texttt{volatility\_12}
& $\operatorname{std}(RET_{t-11:t})$
& Trailing 12 months\par CRSP: RET \\

\texttt{parkinson\_vol}
& $\displaystyle
\sqrt{
\frac{1}{4\ln 2}
\operatorname{mean}_{\tau=t-11,\ldots,t}
\left[
\ln^2\!\left(
\frac{\mathrm{adjhi}_{\tau}}
{\mathrm{adjlo}_{\tau}}
\right)
\right]
}$
& Trailing 12 months\par CRSP: ASKHI, BIDLO, CFACPR \\[5pt]

\texttt{bb\_width}
& $\displaystyle
\frac{4\,\operatorname{std}(adjp_{t-11:t})}{ma12_t}$
& Trailing 12 months\par CRSP: PRC, CFACPR \\[5pt]

\texttt{atr}
&
\(
\begin{aligned}
\mathrm{TR}_t
&= \max\{\mathrm{adjhi}_t-\mathrm{adjlo}_t,\\
&\qquad |\mathrm{adjhi}_t-\mathrm{adjp}_{t-1}|,\\
&\qquad |\mathrm{adjlo}_t-\mathrm{adjp}_{t-1}|\},\\
\mathrm{ATR}_t
&= \mathrm{EMA}_{14}(\mathrm{TR}_t)
\end{aligned}
\)
&
14-month EMA\newline
CRSP: ASKHI, BIDLO,\newline
PRC, CFACPR
\\[10pt]

\texttt{beta60m}
& $\displaystyle
\beta_{i,t}=
\frac{
\operatorname{Cov}_{\tau=t-60,\ldots,t-1}
(RET_{i,\tau},VWRETD_{\tau})
}{
\operatorname{Var}_{\tau=t-60,\ldots,t-1}
(VWRETD_{\tau})
}$
& Trailing 60 months\par CRSP: RET, VWRETD \\

\texttt{retskew12m}
& $\displaystyle
\operatorname{skew}\!\left(RET_{t-12:t-1}\right)$
& Previous 12 months\par CRSP: RET \\

\texttt{maxret12m}
& $\displaystyle
\max\!\left(RET_{t-12:t-1}\right)$
& Previous 12 months\par CRSP: RET \\

\addlinespace[3pt]


\multicolumn{3}{@{}l}{\textbf{Volume \& Liquidity}}\\
\addlinespace[2pt]

\texttt{volume}
& $VOL_t$
& Month $t$\par CRSP: VOL \\

\texttt{volume\_ma12}
& $\operatorname{mean}(VOL_{t-11:t})$
& Trailing 12 months\par CRSP: VOL \\

\texttt{volume\_ratio}
& $\displaystyle \frac{VOL_t}{volume\_ma12_t}$
& Month $t$, relative to 12M mean\par CRSP: VOL \\[5pt]

\texttt{obv\_normalized}
&
\(
\begin{aligned}
OBV_t
&= \sum_{\tau\leq t}
\operatorname{sign}(RET_\tau)\,VOL_\tau,\\
OBV^{norm}_t
&=
\frac{
OBV_t-\operatorname{ExpMean}(OBV)_t
}{
\operatorname{ExpStd}(OBV)_t+10^{-12}
}.
\end{aligned}
\)
& Expanding window\par CRSP: RET, VOL \\[5pt]

\texttt{turnover\_ratio}
& $\displaystyle \frac{VOL_t}{SHROUT_t}$
& Month $t$\par CRSP: VOL, SHROUT \\

\texttt{dolvol}
& $|PRC_t|\times VOL_t$
& Month $t$\par CRSP: PRC, VOL \\

\texttt{amihud}
& $\displaystyle \frac{|RET_t|}{dolvol_t}$
& Month $t$\par CRSP: RET, PRC, VOL \\

\texttt{bidask}
& $\displaystyle
\frac{ASK_t-BID_t}{(ASK_t+BID_t)/2}$
& Month-end $t$\par CRSP: BID, ASK \\

\end{longtable}

\endgroup

\end{document}